\documentclass[runningheads]{llncs}

\usepackage[mobile]{eccv}

\usepackage{eccvabbrv}

\usepackage{graphicx}
\usepackage{booktabs}

\usepackage[accsupp]{axessibility}  

\usepackage{algorithm}
\usepackage{algpseudocode}
\usepackage{multirow}
\usepackage{pifont}

\algrenewcommand{\Return}{\State\algorithmicreturn~}

\usepackage[pagebackref,breaklinks,colorlinks,citecolor=eccvblue]{hyperref}

\usepackage{orcidlink}
\usepackage{ulem}
\usepackage{booktabs}
\usepackage{multirow}
\usepackage{graphicx}
\usepackage{caption}
\usepackage{soul}
\usepackage{amssymb} 
\usepackage{pifont} 
\definecolor{mygreen}{RGB}{0,128,0} 

\newcommand{\ours}[0]{{HYDRA}}
\newcommand{\memory}[0]{{State Memory Bank}}
\newcommand{\convertor}[0]{{textualizer}}
\newcommand{\sampling}[1]{{Sampling}}

\DeclareMathOperator*{\argmax}{argmax}
\usepackage{listings}
\usepackage{tcolorbox}
\tcbuselibrary{breakable}
\newtcolorbox[list inside=prompt,auto counter,number within=section]{prompt}[1][]{
    colbacktitle=black!60,
    coltitle=white,
    fontupper=\footnotesize,
    boxsep=5pt,
    left=0pt,
    right=0pt,
    top=0pt,
    bottom=0pt,
    boxrule=1pt,
    title={#1},
    #1, 
    breakable,
}
\newtcolorbox[list inside=prompt,auto counter,number within=section]{template}[1][]{
    colbacktitle=black!60,
    coltitle=white,
    colback=yellow!20!white,
    fontupper=\footnotesize,
    boxsep=5pt,
    left=0pt,
    right=0pt,
    top=0pt,
    bottom=0pt,
    boxrule=1pt,
    title={#1},
    #1, 
    breakable,
}
\begin{document}

\title{\includegraphics[height=12.5pt,width=10pt,trim=0 4mm 0 -4mm]{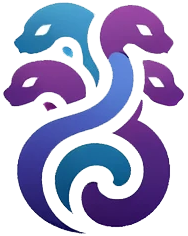} \ours{}: A Hyper Agent for Dynamic Compositional Visual Reasoning} 

\titlerunning{HYDRA}


\author{Fucai Ke$^*$\inst{1,2}\orcidlink{0000-0001-9709-1305}
\and Zhixi Cai$^*$\inst{2}\orcidlink{0000-0001-7978-0860}
\and Simindokht Jahangard$^*$\inst{2}\orcidlink{0000-0002-8903-5175}
\and \\Weiqing Wang\inst{2}\orcidlink{0000-0002-9578-819X}
\and Pari Delir Haghighi\inst{2}\orcidlink{0000-0001-9922-1214}
\and Hamid Rezatofighi\inst{2}\orcidlink{0000-0002-8659-8773}
}

\authorrunning{F.~Ke et al.}

\institute{Building 4.0 CRC, Melbourne 3145, Australia \and Faculty of Information Technology, Monash University, Melbourne 3800, Australia 
\email{\{fucai.ke1,zhixi.cai,simindokht.jahangard,teresa.wang,\\pari.delir.haghighi,hamid.rezatofighi\}@monash.edu}\\
\url{https://hydra-vl4ai.github.io/}
}

\maketitle
\def\thefootnote{*}\footnotetext{These authors contributed equally to this work}

\begin{abstract}
Recent advances in visual reasoning (VR), particularly with the aid of Large Vision-Language Models (VLMs), show promise but require access to large-scale datasets and face challenges such as high computational costs and limited generalization capabilities. Compositional visual reasoning approaches have emerged as effective strategies; however, they heavily rely on the commonsense knowledge encoded in Large Language Models (LLMs) to perform planning, reasoning, or both, without considering the effect of their decisions on the visual reasoning process, which can lead to errors or failed procedures. To address these challenges, we introduce \ours{}, a multi-stage dynamic compositional visual reasoning framework designed for reliable and incrementally progressive general reasoning. \ours{} integrates three essential modules: a planner, a Reinforcement Learning (RL) agent serving as a cognitive controller, and a reasoner. The planner and reasoner modules utilize an LLM to generate instruction samples and executable code from the selected instruction, respectively, while the RL agent dynamically interacts with these modules, making high-level decisions on selection of the best instruction sample given information from the historical state stored through a feedback loop. This adaptable design enables \ours{} to adjust its actions based on previous feedback received during the reasoning process, leading to more reliable reasoning outputs and ultimately enhancing its overall effectiveness. Our framework demonstrates state-of-the-art performance in various VR tasks on four different widely-used datasets.

\end{abstract}    
\section{Introduction}
\label{sec:intro}

Visual reasoning (VR) involves constructing a detailed representation of a visual scene and reasoning through it in steps, similar to human cognition, often in response to textual queries or prompts~\cite{amizadeh2020neuro}. It encompasses various tasks, including but not limited to Visual Question Answering (VQA)~\cite{amizadeh2020neuro}, Visual Commonsense Reasoning (VCR)~\cite{zellers2019recognition}, and Visual Grounding (VG)~\cite{yu2016modeling}.
\begin{figure}[htbp]
\begin{center}
\scalebox{0.96}{
  \includegraphics[width=1\linewidth]{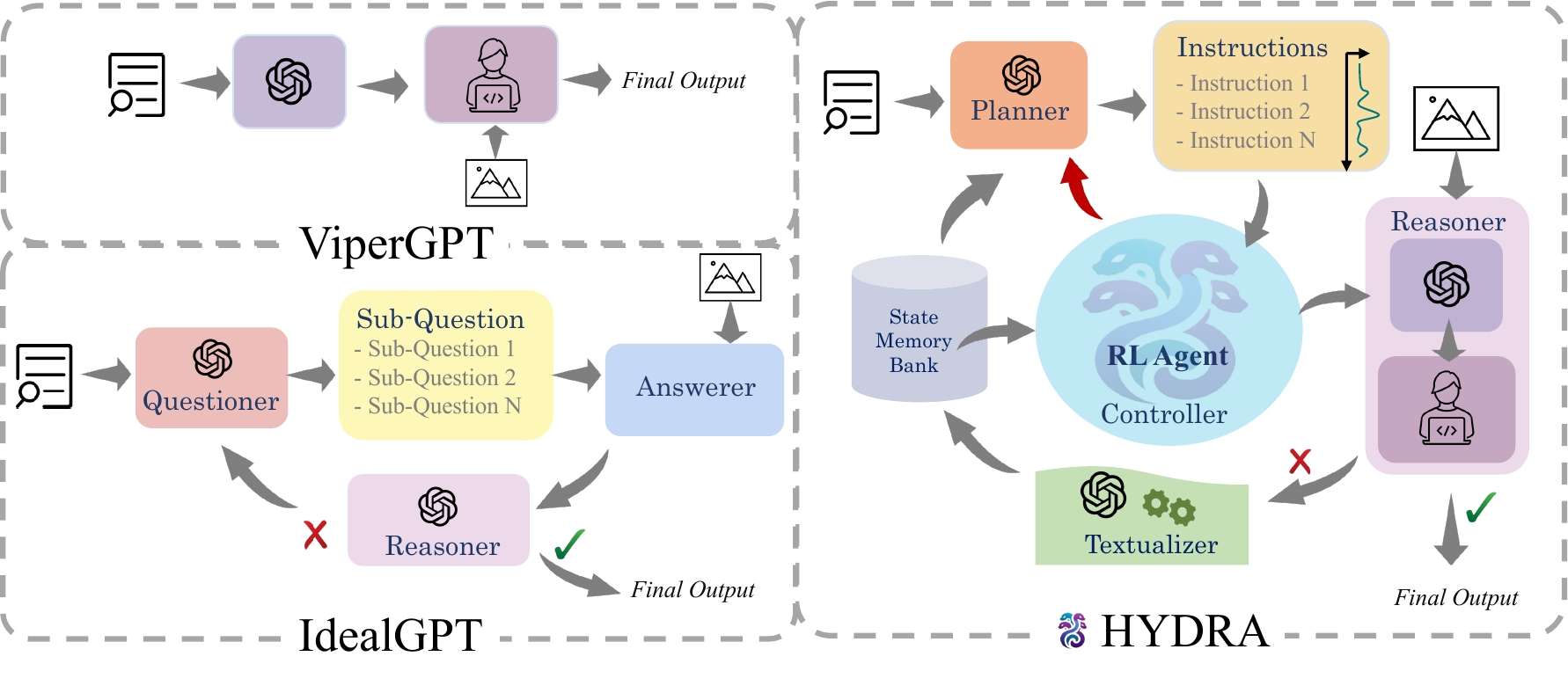} 
}
\end{center}
\caption{Comparison of ViperGPT~\cite{suris2023vipergpt}, IdealGPT~\cite{you2023idealgpt}, and HYDRA: ViperGPT employs a single feed-forward process approach, IdealGPT breaks down questions into sub-questions using a loop, while HYDRA utilizes diverse instructions and an RL agent in an incremental loop for feedback, showcasing its superior adaptability and efficiency in handling complex visual reasoning challenges.}
\label{fig:Teaser}
\end{figure}
In recent years, advancements in Large Language Models (LLMs) ~\cite{chowdhery2022palm,bai2022constitutional,ouyang2022training,radford2019language} and their derivatives, such as VLMs ~\cite{li2023otter,wu2023next,stanic2024towards,zhu2023minigpt} have sparked hope for their effectiveness in solving visual reasoning tasks.
 While these models have shown promising results in certain tasks like VQA and VCR~\cite{zhang2023video}, their training as single monolithic end-to-end models necessitates large-scale datasets, imposing significant computational resource requirements. Additionally, while these models excel within their training domain, they may require further adaptation to achieve reliable performance when applied to diverse datasets or domains~\cite{you2023idealgpt,suris2023vipergpt,stanic2024towards}.\\
In recent advancements, compositional approaches \cite{gupta2023visual, suris2023vipergpt, lu2023chameleon, wu2023visual} have emerged as effective strategies for addressing VR challenges. These approaches break down complex tasks into simpler sub-components, employing a divide-and-conquer methodology. They employ LLMs alongside Visual Foundation Models (VFMs) without requiring extensive training. LLMs can function as planners, code generators, or reasoner, while VFMs act as visual perception components, facilitating structured analysis and task-specific plan generation to enhance adaptability and improve generalization across diverse scenarios. A recent SoTA compositional model is ViperGPT~\cite{suris2023vipergpt}, which utilizes LLMs to generate code programs for visual queries and solve the task in a single feed forward process. 
 IdealGPT~\cite{you2023idealgpt} proposed an enhanced framework by utilising LLMs as both questioners and reasoners, with a pre-trained Vision-Language Model (VLM) serving as the answerer, Figure~\ref{fig:Teaser}. In this model, LLM decomposes main questions into sub-questions, with the reasoner determining whether further sub-question generation is required through iterations or if the final output has been reached.\\
However, these models come with certain limitations. Primarily, the outputs generated by LLMs may sometimes lack meaning, and  when these outputs proceed to subsequent steps without verification, they can impact the outputs of other components, thus adversely affecting overall performance. Moreover, LLMs utilized in the planner or questioner during the initial step lack information from visual content (perception module) in later states to adjust their outputs \cite{gupta2023visual, hu2023visual}. Additionally, the process of generating subsequent questions often begins from scratch without storing information from previous steps, potentially leading to more iterations. Furthermore, these approach heavily rely on commomsense knowledge encoded in LLMs to do planning and reasoning for VR tasks.\\
In this paper, we present \ours{}, a HYper agent for Dynamic compositional visual ReAsoning, an innovative framework designed to address the aforementioned challenges. \ours{} is composed of three main modules planner, controller (Reinforcement Learning-based agent (RL)) and reasoner.  Notably, in the planner, upon receiving textual queries, unlike prior compositional approaches, we employed LLM to generate some instruction samples with varying depths based on a distribution, instead of relying on a single instruction sample. Furthermore, we integrate a hyper RL agent to dynamically interact with some modules to make an high-level decision on the instruction samples generated by LLM in the planner to evaluate their validity. If the RL agent detects any invalid instruction samples, a request is sent back to the planner for alternative suggestions. Conversely, if the instruction samples are considered valid, the chosen instruction sample is forwarded to the reasoner. In the reasoner, the selected instruction sample undergoes analysis by LLM, and the resulting tailored code is sent to the code generator. The code generator employs Python API code to utilize VFMs for additional visual content processing. If the reasoner output is incomplete or fails, the output is converted to textual format in the \convertor module and then stored in \memory. Afterwards, another request is then sent back to the planner to generate new instructions, which are again fed to the controller module to select an instruction sample. This iterative process continues incrementally until the final desired output is achieved.
The design of \ours{} integrates not only the incremental storage of information from previous states (incremental reasoning), considered by the RL agent, but also the capability to utilize feedback from VFMs acquired from earlier perception processes. This enables dynamic adjustment of actions and responses based on feedback from visual perception modules. This innovative design facilitates hyper decision-making by the hyper RL agent, thereby refining reasoning capabilities and overall effectiveness. The overall design of \ours{} compared with the previous compositional approach is shown in Figure~\ref{fig:Teaser}.
 We evaluated our framework on several popular VR datasets and compared it with the advanced models, showing state-of-the-art performance.
In sum, the key contributions of this work are as follows:
\begin{enumerate}
    \item Integrating a cognitive reinforcement learning-based agent as a controller into a framework to foster hyper decision-making and behavior across diverse environments, enhancing system cohesion, performance, and reasoning capabilities.

    \item Employing LLM as a natural language planner that enables the dynamic generation of valid instruction samples for iterative processing. The samples are vary in both the complexity and scope of perception tasks assigned with validity probabilities.
    
    \item Applying incremental reasoning, storing information from previous states aids both the LLM and RL agent in acquiring fine-grained visual information through VFMs and the visual-perception-to-text module, thereby refining their reasoning processes.

\end{enumerate}

\section{Related Work}
\textbf{Single Monolithic End-to-End Methods.}
Recent advancements in Large Language Models (LLMs)~\cite{chowdhery2022palm,bai2022constitutional,ouyang2022training,radford2019language} have notably improved their ability to understand and reason visual content.  Their derivatives, VLMs, like Video-LLaMA \cite{zhang2023video} and NExT-GPT \cite{wu2023next} excel in comprehending detailed videos and seamlessly integrating text, images, videos, and audio for cross-modal reasoning. Otter~\cite{li2023otter}, Flamingo~\cite{alayrac2022flamingo}, and Visual ChatGPT~\cite{wu2023visual} further enhance visual reasoning by integrating visual inputs into their language understanding processes, enabling contextually relevant responses. Initiatives like InstructBLIP~\cite{instructblip}, M$^{3}$IT~\cite{li2023m}, and VisionLLM~\cite{wang2023visionllm} emphasize instruction tuning, multilingual datasets, and vision-centric tasks, advancing language understanding and nuanced video comprehension through a blend of language and visual cues. These developments signal a significant shift towards AI systems proficient in reasoning across textual and visual domains.
However, these single monolithic end-to-end models suffer from reduced interpretability, require significant computational power and extensive training data resources. Besides, these models exhibit limited generalization capabilities due to the vast scale of the trained neural networks~\cite{stanic2024towards}. Various vision challenges often necessitate distinct models, typically involving the manual selection and assembly of specific models tailored to each particular scenario. Given the exponentially large long tail of compositional tasks, the proposed data-intensive and compute-intensive single monolithic end-to-end models may fall short in solving these types of tasks~\cite{villalobos2022will, yang2023mm}.
\begin{table*}[b]
\centering
\caption{Summary of compositional models, including \ours. \textit{IR: Incremental Reasoning. VQA: Visual Question Answering. VG: Visual Grounding. HF: HuggingFace.}}
\label{tab:performance}
{\fontsize{8}{10}\selectfont 
\begin{tabular}{lccccccc}
\toprule
Model & \multicolumn{5}{c}{Module} & \multicolumn{2}{c}{Task} \\
\cmidrule(lr){2-6} \cmidrule(lr){7-8} 
 & Planner & Perception & Reasoner & Controller & IR & VQA & VG\\
Visprog~\cite{gupta2023visual} & \ding{55} & VFMs & GPT-3 & \ding{55} & \ding{55} & \ding{51} &\ding{51} \\
Chameleon~\cite{lu2023chameleon} & ChatGPT & VFMs & ChatGPT & \ding{55} & \ding{55} & \ding{51} & \ding{55}\\
IdealGPT~\cite{you2023idealgpt} & ChatGPT & BLIP2 & ChatGPT & \ding{55}& \ding{55} & \ding{51} & \ding{55} \\
HuggingGPT~\cite{shen2023hugginggpt} & ChatGPT & HF-VFMs & ChatGPT & \ding{55} & \ding{55} & \ding{51} & \ding{51} \\
ViperGPT~\cite{suris2023vipergpt} & \ding{55} & VFMs & Codex & \ding{55} & \ding{55} & \ding{51} & \ding{51}  \\ \hline
\bf{\ours{}} & ChatGPT & VFMs & ChatGPT & RL-Agent & \ding{51} & \ding{51} & \ding{51} \\
\bottomrule
\end{tabular}}
\end{table*}
 Consequently, compositional reasoning, generalization, fine-grained spatial reasoning abilities, and counting capabilities remain significant challenges for even the most advanced, large-scale single monolithic end-to-end models~\cite{bugliarello2023measuring, hsieh2023sugarcrepe,yuksekgonul2022and,suris2023vipergpt}.\\
\textbf{Compositional Visual Reasoning Methods.}
The compositional approach introduces a strategy aimed at addressing the challenges faced by end-to-end VLMs~\cite{suris2023vipergpt, you2023idealgpt, lu2023chameleon, gupta2023visual, stanic2024towards}. 
These models tackle complex tasks by breaking them down into multiple subtasks, solving each one individually, and then utilizing the intermediate outcomes to address the overarching task. These models utilize the potent chain-of-thought (CoT) functionality of LLMs acting as planners, reasoner, etc. This capability facilitates the breakdown of intricate problems into manageable and individually solvable intermediate steps through the provision of instructions \cite{brown2020language, chowdhery2023palm, kojima2022large, huang2022language}. The instructions may take the form of Python execution code that embodies logical operations \cite{gupta2023visual, suris2023vipergpt}. For example, Visprog~\cite{gupta2023visual} and ViperGPT \cite{suris2023vipergpt} seek to eliminate the requirement for task-specific training in both programming logic and perception modules by employing code generation models. These strategies facilitate the assembly of VLMs into subroutines, thereby enabling the production of results. An alternative strategy, emblematic of the divide-and-conquer methodology, is exemplified by IdealGPT~\cite{you2023idealgpt}. This approach harnesses a captioning model for the acquisition of elementary visual data and engages a LLM to serve as a planner. The high-level inquiries are methodically deconstructed into three distinct sub-questions, which are processed concurrently. Following this, perception tools (VFMs) are employed to individually address each sub-question. The outcomes are then aggregated and analyzed by the reasoning mechanism to deduce the comprehensive final response. Moreover, the activation status and the sequential order of VFMs, as utilized by visual perception tools, constitute a form of instruction \cite{lu2023chameleon}. The system implements predefined functionalities based on these instructions to systematically activate perception tools in a specified sequence. This process culminates in the aggregation of data, which is subsequently analyzed by the reasoning mechanism to formulate the ultimate conclusion.\\
All these compositional processing heavily depends on the capability of LLMs to perform commonsense reasoning and make decisions. 
However, despite their capabilities, LLMs have certain limitations. Primarily, the outputs they generate may lack meaningfulness, and if these outputs proceed to subsequent steps without verification, they can adversely affect the performance of other components. Additionally, LLMs used in planning or questioning lack access to visual content information in later stages, which hinders their ability to adjust outputs accordingly. Moreover, the process of generating subsequent questions often starts anew without retaining information from previous steps, potentially leading to more iterations. Furthermore, these methodologies heavily rely on the common-sense knowledge encoded in LLMs for planning and reasoning within virtual reality tasks.
In this paper, we introduce a new framework that utilizes a cognitive reinforcement learning-based agent to address these challenges. This framework enhances decision-making, system performance, and reasoning across different scenarios. Moreover, we effectively harness LLM knowledge to generate instructional samples and facilitate incremental reasoning for acquiring detailed visual information. A comparison between recent compositional VR models and our approach is presented in Table~\ref{tab:performance}.




\section{Approach} \label{sec:approach}
\begin{figure*}[t]
\begin{center}
\scalebox{0.95}{
  \includegraphics[width=1\linewidth]{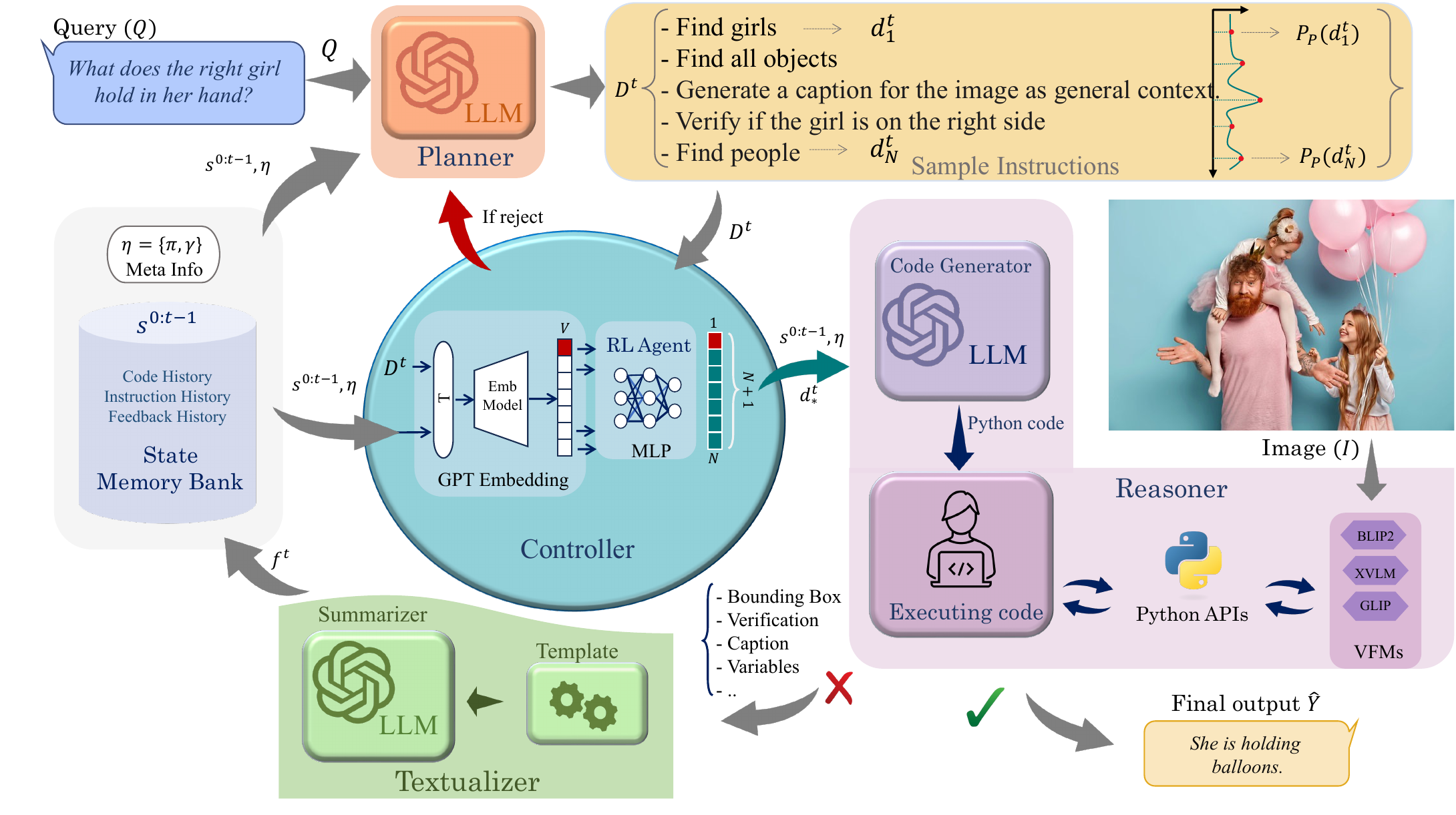}
}
\end{center}
\caption{
The \ours{} detailed design includes key modules: planner, controller, reasoner, \convertor{}, \memory{}~($s^{t-1}$), and meta information ($\eta$). Input Q is given to the planner to generate instructions $D^t$ using $s^{t-1}$ and $\eta$. The controller receives $D^t$, and if invalid, requests alternative samples from the planner. Otherwise, it sends chosen instruction $d^t_*$ to the reasoner, which generates perceptual output using Python APIs and VFMs. Incomplete output is converted to textual format, $f^t$, by the textualizer and stored in \memory{}. This process iterates until the desired final output, $\hat{Y}$, is achieved.}
\label{fig:framework}
\end{figure*}
\begin{figure}[tb]
\begin{algorithm}[H]
\caption{\ours{} Inference}
\label{alg:inference}
\begin{algorithmic}[1]
\Require{$X,\mathcal{F}_P,\mathcal{F}_C,\mathcal{F}_R,\mathcal{F}_T,\eta,\theta$}
\State $\{Q, I\} \gets X; t \gets 1; f \gets \{\};d \gets \{\}$ \Comment{Initialize the inputs and state}
\While{not final}
    \State $s^{0:t-1} \gets \{f,d\}$
    \State $D^t \gets\mathcal{F}_P(Q, s, \eta)$ \Comment{Generate instructions}
    \State $d^t_* \gets \operatorname*\argmax_{d_i^t \in D^t}\mathcal{F}_C^{\theta}(D^t, s^{0:t-1}, \eta) * P_P(d_i^t) $ \Comment{Select the optimal instruction}
    \If{$D^t$ is rejected}
        \textbf{go to} 4
    \EndIf
    \State $f^t \gets \mathcal{F}_T(\mathcal{F}_R(Q, d^t_*, s, \eta))$ \Comment{Execute code and textualize perception results}
    \If{execution error}
        \textbf{go to} 8
    \EndIf
    \State $t \gets t + 1$
    \State $f.\text{append}(f^t);d.\text{append}(d_*^t)$ \Comment{Update the state}
\EndWhile
\State $\hat{Y} \gets \text{Extract answer from}\ f$ \Comment{Resolve the final answer}\\
\textbf{return} $\hat{Y}$
\end{algorithmic}
\end{algorithm}
\end{figure}
The design of \ours{} are provided Figure~\ref{fig:framework} in detail, comprising several key modules: planner ($\mathcal{F}_P$), controller ($\mathcal{F}_C^{ \theta}$), reasoner($\mathcal{F}_R$), \convertor{} ($\mathcal{F}_T$), a \memory{} and meta information ($\eta$).
The framework's input comprises query-image pairs, denoted as $X = \{Q,I\}$, and the final output, $\hat{Y}$, can be textual answers or bounding boxes for the visual grounding task.
The planner $\mathcal{F}_P$, utilizing LLM, generates some instruction samples based on the input query $Q$  using some information from meta information and \memory{}. Then, the generated instruction samples are fed to controller $\mathcal{F}_C$ which is composed of GPT embedding and RL agent that evaluate the validity of instruction samples. If the RL agent detects invalid instruction samples, it forwards a request to the planner for alternative instruction samples; conversely, an instruction sample is picked as the chosen sample, $d_*$, and sent to the reasoner. The chosen instruction sample is fed to the LLM in the reasoning module, and the corresponding Python code is generated in the code generator submodule. Subsequently, this Python code is executed in the executing code submodule utilizing Python APIs and VFMs. If the output is incomplete or unsuccessful, it is converted to textual format through the \convertor{} module and stored in the \memory{}. Thereafter, another request is sent to the planner to generate new instruction samples, which are then provided to the controller module to select a valid instruction sample. This iterative process continues incrementally until the desired final output is obtained.\\
As \ours{} is a framework that operates through several iterations to simplify the process, we use $s^{0:t}$ to depict the progression from the initial state to the current state $0:t$. Additionally, in the first iteration, there is no information from the previous iteration, denoted as $s^{0}=\{ \}$. Note that all LLMs in the planner, reasoner, and \convertor{} are the same, with only their prompts being changed in different modules, and for enhanced clarity, we present them separately in the figure. The algorithm of the whole inference process is provided in Algorithm~\ref{alg:inference}.
The technical details for each module, along with further elaboration, are provided in the following. \\
\textbf{\memory~\& Meta Information.} As \ours{} progresses through multiple iterations and considers information from previous ones, all data, including code, instruction, and the output of the reasoner from former iteration, are stored in \memory{}, represented by a grey cylinder in Figure ~\ref{fig:framework}. Furthermore, meta information encompasses crucial data such as a subset of skills $\pi \in \Pi$ and various task descriptions $\gamma \in \Gamma$ tailored for different tasks that the LLM needs as a prompt. For simplicity, these are denoted as $\eta = {\gamma, \pi}$ in the subsequent equations.\\
\textbf{Planner Module.}
Highlighted in \textcolor{orange}{orange} in Figure~\ref{fig:framework}, this module receives $Q$ and other data from the \memory{}. It generates N instruction samples (e.g., "find girls", "verify if the girl is on the right side"), $d_i^{t}$ of varying depth, where each instruction sample can have different actions or levels of complexity. For instance, some instructions may involve simple tasks, while others may entail more intricate actions or multi-step processes. Along with these instruction samples, corresponding confidence probabilities $P_p(d_i^{t})$ are provided, indicating the likelihood of each instruction being accurately executed. These outputs are generated by the LLM ChatGPT\footnote{\label{foot:chatgpt}\url{https://chat.openai.com/}} and are represented by $D^t = \{(d_i^{t}, P_P(d_i^{t}))\}_{i=1}^N$ in the yellow box. This process is described by the equation: 
\begin{equation}
D^t = \mathcal{F}_P(Q, s^{0:t-1},\eta)
\end{equation}
\noindent\textbf{Controller Module.}
This module serves as the central component of \ours{}, dynamically interacting with other modules to facilitate hyper decision-making and functioning as a cognitive controller.This module integrates embedding, leveraging GPT-3~\cite{brown2020language}, to extract the features highlighted in a cyan  circle in the Figure~\ref{fig:framework}. It takes $D^t$, $\eta$ and $s^{0:t-1}$ and embeds them into a vector, $V$. Subsequently, it passes through an RL agent, which consists of a trainable MLP layer followed by a softmax function with an output size of N+1.
Through this module, the instruction samples undergo evaluation and if the RL agent considers them invalid, a request is sent to the planner to regenerate new instruction samples, as indicated by the \textcolor{red}{red arrow} in Figure~\ref{fig:framework}. Otherwise, the chosen instruction sample, $d_*^t$, is selected and proceeds to the reasoner, depicted by the {\color{ForestGreen} green arrow}.
\begin{align}
d^t_* 
&= \argmax_{d_i^t \in D^t}\mathcal{F}_C^{\theta}(D^t, s^{0:t-1}, \eta) * P_P(d_i^t) \
\end{align}
\noindent\uline{\textbf{Training phase.}} As mentioned earlier, the RL agent is a trainable MLP layer based on Reinforcement Learning, employing the DQN algorithm~\cite{mnih2013playing}. During the training phase, the objective of the RL agent is to maximize the expected cumulative reward. The reward function is designed to favour fewer iterations and correct output while penalizing more iterations and incorrect output. We iteratively accumulate the reward function as shown below.
\begin{equation} \label{Reward}
    R^{t}= \begin{cases}
    R^{t-1} - t  & \text{if not final step},\\
    R^{t-1} + \alpha m & \text{if answer is relate} \\
    R^{t-1} - \alpha & \text{if answer is unrelated} \\
    R^1 & \text{if $t = 1$}
\end{cases}
\end{equation}\\
where $m$ is the performance metrics (e.g. accuracy, intersection over union), $\alpha$ and $R^1$ are the hyperparameter constants. Additional details regarding this phase are provided in the supplementary material.\\
\noindent\textbf{Reasoner Module.} 
Illustrated in {\color{Rhodamine}light pink} in the Figure~\ref{fig:framework}, this module consists of an LLM as code generator and a code executor sub-module. In this setup, ChatGPT\footref{foot:chatgpt} receives the selected instruction sample \(d^t_*\) from the controller module, along with necessary information from the previous iteration, $s^{0:t-1}$, and  $\eta$ , to generate Python code. This Python code is then transferred to the execution sub-module within perception tools, such as VFMs including GLIP~\cite{li2022grounded}, BLIP2~\cite{li2023blip}, LLaVA-1.5~\cite{liu2024improved}, MiDaS~\cite{ranftl2020towards}, and XVLM~\cite{zeng2021multi}.
Python interpreter to execute the code in the Python context loaded with the predefined Python APIs. In the execution, all the variable values (perceptual output) are collected and logged and will be sent to the next module via the feedback. 
\begin{equation}
  \text{ perceptual output} = \mathcal{F}_R(Q, d^t_*, s^{0:t-1}, \eta)
\end{equation}
\textbf{Textualizer Module.}
If the perceptual output from the reasoner module is incomplete or unsuccessful, it undergoes conversion to textual format within this module, as depicted by the \textcolor{ForestGreen}{green} in Figure~\ref{fig:framework}. The perceptual output from the reasoner, which may consist of bounding boxes, verifications, or captions, is transformed into a textual format using a template. This conversion ensures that the input is understandable for the LLM and ensures that all information stored in the \memory~has the same format that can be used in the next iterations. Subsequently, the LLM summarizes the current state information, $f^t$, and stores it in \memory. Further details about these templates are available in the supplementary material.
\\
\noindent\textbf{Technical Details:}
The iterative process continues incrementally until the desired final output is achieved, which we refer to as the incremental reasoning mechanism. It's worth noting that the \ours{} does not always require iterations; by efficiently integrating the RL agent, the final output of the task can be generated in just a single iteration. That could be due to the simplicity of the task, or the RL agent may choose to select an instruction that includes all the necessary steps for generating the final output in a single iteration.
\section{Experiments and Results}
\label{sec:experiments}
\textbf{Implementation Details:}
To train our framework, we utilized PyTorch~\cite{Paszke_Gross_Massa_Lerer_Bradbury_Chanan_Killeen_Lin_Gimelshein_Antiga_etal._2019} with NVIDIA RTX 4090 GPUs, employing a learning rate of $1 \times 10^{-4}$ and a batch size of 128. The Multi-Layer Perceptron (MLP) used for the RL agent, consists of three layers with dimensions 1536, 512, and 6. The hyper-parameters for reinforcement learning are set as $R^1 = 100$ and $\alpha = 100$. During the training process, early stopping is applied once the reward converges.
For a fair comparison, we evaluated the state-of-the-art (SoTA) baselines using configurations from their official code repositories and papers~\cite{stanic2024towards}. We utilized the largest available backbone for the end-to-end VLMs. We also replaced ChatGPT as the code generator in ViperGPT~\cite{suris2023vipergpt}, given the discontinuation of GPT3 Codex by OpenAI~\cite{suris2023vipergpt}.
Supplementary materials provide additional details on implementation including instructions and prompts for the planner, code generator, and controller.\\
\textbf{Datasets and Evaluation Metric:} 
We evaluated our framework across three key tasks in visual reasoning. Firstly, External Knowledge-dependent Image Question Answering, for which we utilize the OK-VQA dataset~\cite{Marino_Rastegari_Farhadi_Mottaghi_2019} and evaluate performance based on accuracy (ACC) score~\cite{suris2023vipergpt, you2023idealgpt}. Secondly, Compositional Image Question Answering, where the GQA~\cite{hudson2019gqa} dataset serves as our benchmark, again measured by ACC score~\cite{suris2023vipergpt, you2023idealgpt}. Lastly, Visual Grounding tasks are addressed using the RefCOCO~\cite{Yu_Poirson_Yang_Berg_Berg_2016} and RefCOCO+~\cite{Yu_Poirson_Yang_Berg_Berg_2016} datasets, with evaluation based on Intersection over Union (IoU) metrics~\cite{heigold2023video, minderer2024scaling, li2022grounded, subramanian2022reclip, Peng_Wang_Dong_Hao_Huang_Ma_Wei_2023}. These diverse tasks and corresponding datasets offer comprehensive assessments, collectively contributing to the advancement of our framework's capabilities in visual understanding and interpretation.\\
\textbf{Visual Reasoning Tasks and Result Analysis:}
Detailed elaboration and both quantitative and quantitative results for each task, External Knowledge-dependent Image Question Answering, Visual Grounding, and Compositional Image Question Answering, respectively, are provided below.\\
\noindent\uline{\textbf{External Knowledge-dependent Image Question Answering}}~involves using external sources of information, such as databases, to provide context and answer questions about images that cannot be inferred solely from visual content~\cite{suris2023vipergpt}. Following previous works~\cite{suris2023vipergpt}, we additionally employ the LLM~\cite{brown2020language} knowledge with the module llm-query. 
The quantitative results from Table~\ref{subtab:okvqa} highlight the comparison between end-to-end models and compositional models, including \ours{}, on the OK-VQA dataset. \ours{} surpasses previous models by 48.6\%, showcasing a remarkable improvement. 
The incorporation of advanced techniques in \ours{}, such as incremental reasoning mechanisms and leveraging LLM for generating different instructions, greatly contributes to its outstanding performance. 
\begin{table}[tb]
\centering
\caption{Performance on External Knowledge-dependent Image Question Answering and Visual Grounding tasks.}
\label{combined_table}
\begin{subtable}[t]{0.49\textwidth}
\centering
\caption{Performance on OK-VQA.}
\label{subtab:okvqa}\scalebox{1}{
{\fontsize{8}{10}\selectfont
\begin{tabular}{c|lcc}
\toprule
Type & Method & ACC(\%)\\
\midrule
\multirow{7}{*}{E2E} & PNP-VQA~\cite{tiong2022plug} & 35.9\\
    & PICa~\cite{yang2022empirical} & 43.3 \\
    & BLIP-2~\cite{li2023blip} & 45.9 \\
    & Flamingo (9B)~\cite{alayrac2022flamingo} & 44.7\\
    & MiniGPT-4 (13B)~\cite{zhu2023minigpt} & 37.5\\
    & LLaVA (13B)~\cite{liu2023visual} & 42.5 \\
    & InstructBLIP (13B)~\cite{instructblip} & 47.9 \\
\midrule
\multirow{3}{*}{Comp} & IdealGPT~\cite{you2023idealgpt} & 19.4\\
     & ViperGPT~\cite{suris2023vipergpt} & 40.7 \\
     & \textbf{\ours} & \textbf{48.6}\\
\bottomrule
\end{tabular}}}
\end{subtable}
\hfill
\begin{subtable}[t]{0.49\textwidth} 
\centering
\caption{Performance on RefCOCO and RefCOCO+}
\label{subtab:vgt}\scalebox{1}{
\begin{tabular}{@{} c|lcc @{}} 
\toprule
& &\multicolumn{2}{c}{IoU(\%)}\\
Type & Method & Ref &  Ref+   \\
\midrule
\multirow{5}{*}{E2E} & OWL-ViT~\cite{heigold2023video} & 30.3 & 29.4\\
    & OWLv2~\cite{minderer2024scaling} & 33.5 & 31.7 \\
    & GLIP~\cite{li2022grounded} & 55.0 & 52.2\\
    & ReCLIP~\cite{subramanian2022reclip} & 58.6 & 60.5 \\
    & KOSMOS-2~\cite{Peng_Wang_Dong_Hao_Huang_Ma_Wei_2023} & 57.4 & 50.7 \\ 
\midrule
\multirow{3}{*}{Comp} & Code-bison~\cite{stanic2024towards} & 44.4 & 38.2 \\
     & ViperGPT~\cite{suris2023vipergpt} & 59.8 & 60.0\\
     & \textbf{\ours} & \textbf{61.7} & \textbf{61.1} \\
\bottomrule
\end{tabular}}
\end{subtable}
\end{table}

\noindent\uline{\textbf{Visual Grounding}} involves predicting bounding boxes based on the input prompt. \ours{} are equipped with reasoner module which contain grounding-related VFM APIs such as find, exists, and verify-property, similar to ViperGPT. Our method, as shown in Table~\ref{subtab:vgt}, surpasses the state-of-the-art baselines for IoU on RefCOCO~\cite{Yu_Poirson_Yang_Berg_Berg_2016} and RefCOCO+\cite{Yu_Poirson_Yang_Berg_Berg_2016} datasets. Among the end-to-end methods, grounding-specialized approaches like GLIP\cite{li2022grounded} and ReCLIP~\cite{subramanian2022reclip} achieve superior performance compared to the VLM KOSMOS-2~\cite{Peng_Wang_Dong_Hao_Huang_Ma_Wei_2023}. Considering that KOSMOS-2 can also handle other text-based tasks. When comparing methods between end-to-end and compositional approaches, we observe that both compositional visual reasoning approaches (ViperGPT~\cite{suris2023vipergpt} and \ours{}) achieve better performance than end-to-end baselines. This indicates that the compositional approach design is more adept at solving the VG task.\\
\noindent\uline{\textbf{Compositional Image Question Answering}} contains complex questions. These questions require the decomposition into simpler steps for answering. 
Similar to previous works~\cite{suris2023vipergpt}, we utilize the BLIP2~\cite{li2023blip} API simple-query to enhance our understanding of image content.
As demonstrated in Table~\ref{subtab:gqa} with implementation on the GQA dataset, among the end-to-end models, the 30.8\% performance of MiniGPT underscores the importance of instruct tuning. IdealGPT surpasses ViperGPT in performance by leveraging a planner to enhance reasoning capability. Notably, ViperGPT's performance is impeded by the generation of non-executable code snippets, while \ours{} enhances code quality through the integration of multiple sampling and a RL agent controller for code validation, leading to superior performance compared to ViperGPT. Additionally, it highlights that \ours{} achieves an impressive accuracy of 47.9\%, underscoring its robustness and effectiveness in handling the GQA dataset. Further results can be found in the supplementary materials.\\

\begin{table}[tb]
\centering
\caption{Performance on GQA Dataset}
\label{subtab:gqa}
\scalebox{0.9}{\begin{tabular}{clc}
\toprule
Type & Method & ACC(\%) \\ 
\midrule
\multirow{5}{*}{E2E} & BLIP-2~\cite{li2023blip} & 45.5 \\
    & MiniGPT-4 (13B)~\cite{zhu2023minigpt} & 30.8\\
    & LLaVA (13B)~\cite{liu2023visual} & 41.3\\
    & PandaGPT (13B)~\cite{su2023pandagpt} & 41.6\\
    & ImageBind-LLM (7B)~\cite{han2023imagebind} & 41.2\\
\midrule
\multirow{3}{*}{Comp} & IdealGPT~\cite{you2023idealgpt} & 41.7 \\
     & ViperGPT~\cite{suris2023vipergpt} & 37.9 \\
     & \textbf{\ours} & \textbf{47.9} \\
\bottomrule
\end{tabular}}
\end{table}

\begin{table}[tb]
\centering
\caption{Generalization performance for the RL-Agent. The \textit{Train} column is the training data for training the RL agent, and the \textit{Test} column is the test data for evaluating the method.}
\scalebox{0.85}{\begin{tabular}[b]{l|cc|c}
\toprule
Method & Train & Test & ACC(\%) \\
\midrule
ViLT~\cite{pmlr-v139-kim21k} & GQA & OK-VQA & 32.13 \\
ViperGPT~\cite{suris2023vipergpt} & \textbf{--} & OK-VQA & 40.74 \\ \hline
\ours{} & GQA & OK-VQA & 48.17\\
\ours{} & OK-VQA & OK-VQA & 48.63 \\
\hline
\ours{} & OKVQA & A-OKVQA &55.94 \\
\ours{} & A-OKVQA & A-OKVQA &56.35 \\
\bottomrule
\end{tabular}}
\label{generalization}
\vspace{-1em}
\end{table}

\begin{figure}[tb]
\begin{center}
\includegraphics[width=0.85\linewidth]{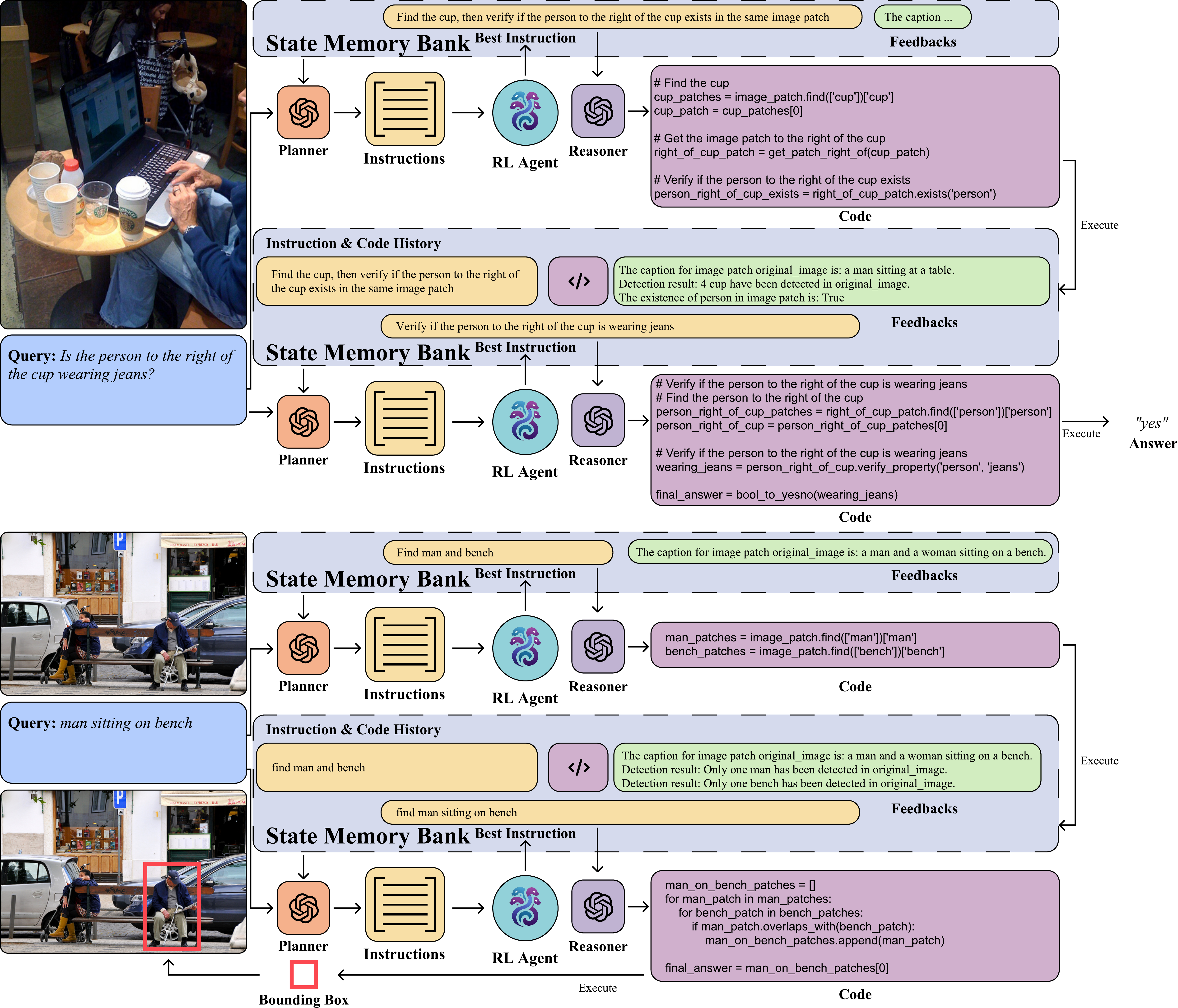} 
\end{center}
\caption{Detailed result examples from \ours{}. The first example describes the intermediate results of the full two iterations in the loop for question answering, whereas the second example is about the grounding task.}
\label{fig:Pic_results}
\end{figure}

\begin{figure}[tb]
\begin{center}
\includegraphics[width=0.8\linewidth]{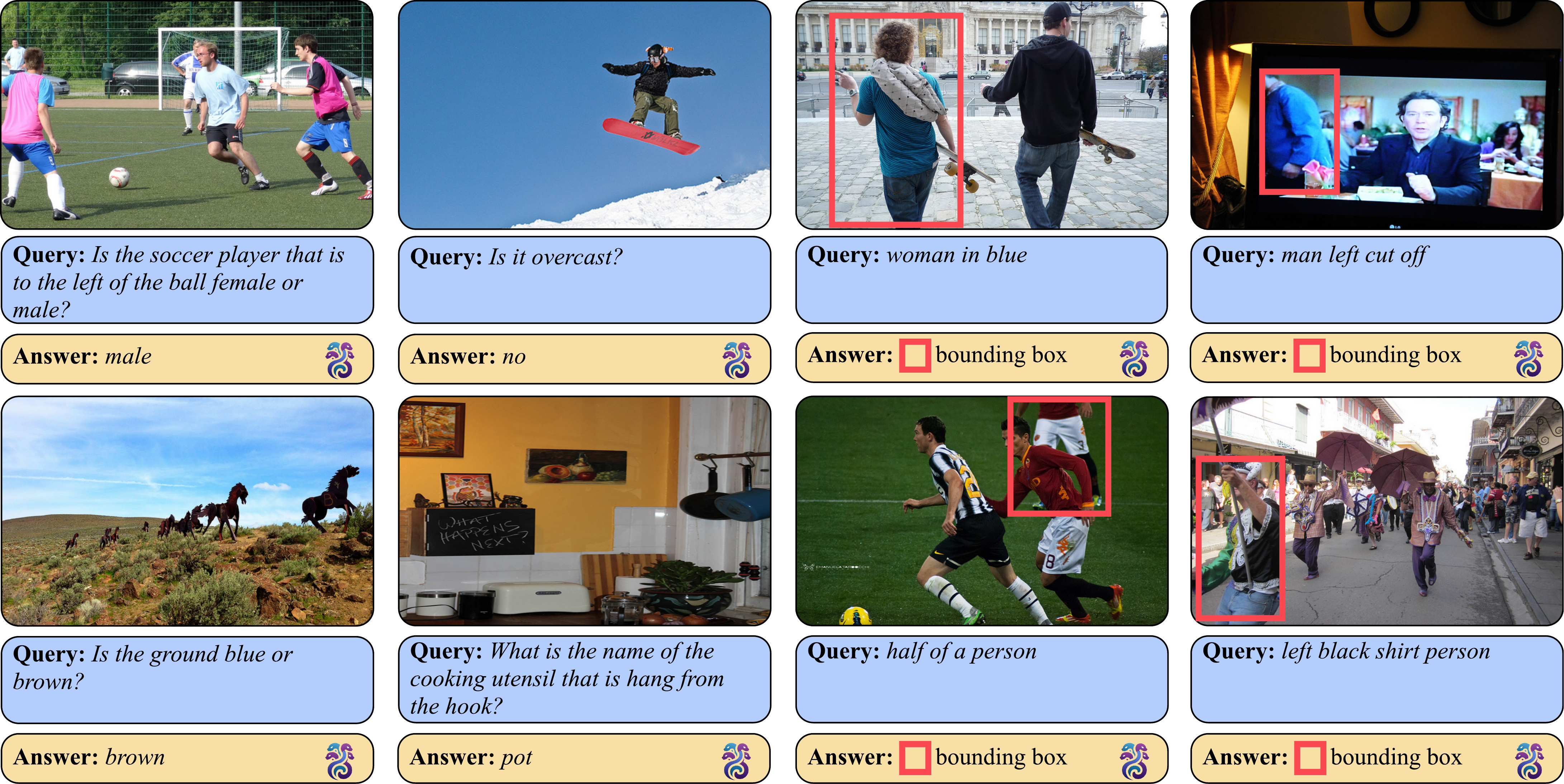} 
\end{center}
\caption{More result examples from \ours{} for question answering and visual grounding tasks.}
\label{fig:Pic_results2}
\end{figure}

\begin{table}[hbt]
\centering
\caption{An ablation study for \ours{} on GQA.}\scriptsize
\scalebox{1}{\begin{tabular}{c|ccc|c}
\toprule
Models& RL-Agent & IR & Sampling & ACC \\
\midrule  
ViperGPT&\ding{55} & \ding{55} & \ding{55} & 37.94 \\
&\ding{51} & \ding{55} & \ding{55} & 43.71  \\ 
&\ding{55} & \ding{55} & \ding{51} & 39.84 \\
&\ding{51} & \ding{55} & \ding{51} & 45.98  \\ 
&\ding{55} & \ding{51} & \ding{55} & 41.08 \\
&\ding{51} & \ding{51} & \ding{55} & 47.07  \\ 
&\ding{55} & \ding{51} & \ding{51} & 46.93 \\
HYDRA&\ding{51} & \ding{51} & \ding{51} & \textbf{47.88} \\
\bottomrule
\end{tabular}}
\label{ablation}
\end{table}

\noindent{\textbf{Generalization Analysis:}
Generalization abilities play a crucial role in adapting approaches to unseen data distributions without necessitating re-training. Given that the RL agent in \ours{} is the sole component requiring training, we conducted generalization experiments on the OK-VQA and A-OK-VQA~\cite{schwenk2022okvqa} dataset, as presented in Table~\ref{generalization}, to assess the module's capacity to operate effectively on unseen data without explicit training. ViLT~\cite{pmlr-v139-kim21k} is chosen as the baseline end-to-end method, which does not require expensive computational resources. 
Notably, the performance of our model, \ours{}, in the cross-dataset experiments (\ie, training on GQA and testing on OK-VQA, and training on OK-VQA and testing on A-OK-VQA) closely matches intra-dataset performance as shown in Table~\ref{generalization}. 
Furthermore, this cross-dataset performance surpasses that of the baseline ViLT~\cite{pmlr-v139-kim21k}, which achieved an accuracy of 32.13\%. Additionally, ViperGPT ~\cite{suris2023vipergpt} exhibits superior performance compared to ViLT, showcasing the superiority of compositional over end-to-end methods in generalizability. 
Comparison with ViperGPT also reveals superior performance, as \ours{} trained on alternative datasets achieved accuracies of 48.17\%. 
These findings underscore the generalizability of the RL agent controller within \ours{}.\\

\noindent\textbf{Qualitative Analysis:} Figure~\ref{fig:Pic_results} demonstrates intermediate processes of \ours{} for two examples, one for visual question-answering and one for visual grounding tasks. We show detailed examples with multiple steps in the first example in each figure, and the brief examples only show the last iteration in the loop. It is observed that the meaningful perception results are summarized as useful feedback for the next iteration of planning and reasoning. Figure~\ref{fig:Pic_results2} includes more qualitative examples of the results using \ours{} on these tasks.

\begin{figure}[tb]
\begin{center}
\includegraphics[width=0.8\linewidth]{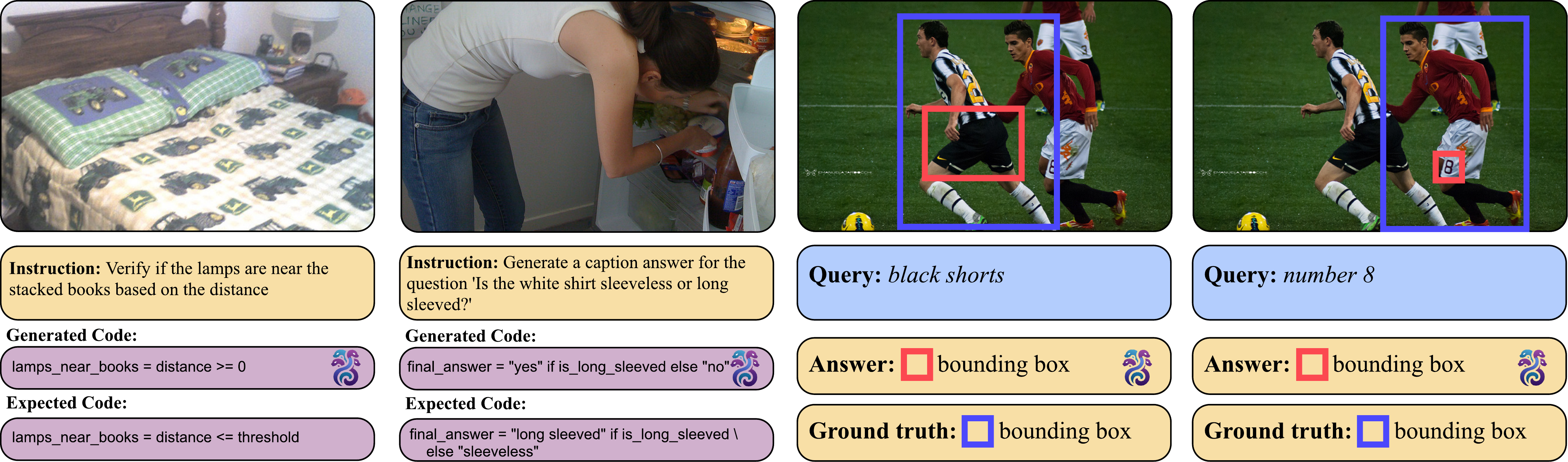} 
\end{center}
\caption{Failure result examples from \ours{}. The left two samples are due to wrong generating codes. The right two failure cases are due to wrong annotation. }
\label{fig:fail}
\end{figure}


\noindent\textbf{Failure Analysis.}
While \ours{} has achieved SoTA performance, there is still room for further improvement in its design. In complex cases, as illustrated in Figure~\ref{fig:fail}, \ours{} may fail due to potential mistakes made by the LLMs within the reasoner and textualizer module. In future iterations, we plan to enhance the complexity of the RL agent, enabling it to exert greater control over the output of LLMs, whether functioning as code generators or summarizers. Additional fail rate analysis can be found in the supplementary materials.
\subsection{Ablation Study}
In this section, we provide an ablation study on suggested key components of \ours{} demonstrating their contributions to the final results.  

\noindent\textbf{Component Analysis.}
As previously mentioned, there are three main contributions in \ours{}: the RL agent, \sampling~ (involving instruction sampling numbers), and Incremental Reasoning (IR). Through this experiment, the efficacy of each component is evaluated and presented in Table~\ref{ablation}. As depicted in Table~\ref{ablation}, the first column displays the models and their variants, while the following three columns represent each key component: RL agent, Incremental Reasoning (IR), and \sampling, respectively. The last column, denoted as ACC, represents the accuracy achieved by each model on the GQA dataset. 
As shown in Table~\ref{ablation}, the RL-Agent significantly improves the overall architecture, achieving an average enhancement of $4.71\%$ in accuracy compared to the variants with the same settings on IR and sampling but without the RL-Agent. Additionally, both IR and \sampling~ further boost the framework's performance by $3.87\%$ and $2.70\%$ on average, compared with the corresponding variant without IR or sampling. Further implementation details can be found in the supplementary.

\section{Conclusion}
In this paper, we introduced \ours{}, a multi-step dynamic compositional visual reasoning framework designed to improve reasoning steadily and reliably. \ours{} combines three key parts: a planner, a RL agent acting as a cognitive controller, and a reasoner. The planner and reasoner modules use an LLM to create instruction samples and executable code from chosen instructions, while the RL agent interacts with these modules to make decisions based on past feedback, adjusting its actions as needed. This flexible setup allows \ours{} to learn from previous experiences during the reasoning process, resulting in more dependable outcomes and overall better performance. In future, our goal is to enhance our framework by fostering greater interaction between the LLM in the reasoner and the texturizer module to mitigate potential errors.

\section*{Acknowledgements}
This research is partially supported by Building 4.0 CRC. We acknowledge the support of the Commonwealth of Australia through the Cooperative Research Centre Programme. Additionally, this material is based on research partially sponsored by the DARPA Assured Neuro Symbolic Learning and Reasoning (ANSR) program under award number FA8750-23-2-1016.


%
%
\bibliographystyle{splncs04}
\bibliography{main}

\clearpage
\appendix
{
\newpage
    \centering
    \Large
    \textbf{\includegraphics[height=12.5pt,width=10pt,trim=0 4mm 0 -4mm]{pic/HYDRA_icon_minimal.png} \ours{}: A Hyper Agent for Dynamic Compositional Visual Reasoning}\\
    \vspace{0.5em}
    Supplementary Material \\
    \vspace{1.0em}
}

\noindent We provide additional details about our framework, HYDRA, as supplementary material. We provide more details about the number of instruction samples, our approach to training RL agents, the templates used in the textualizer module, additional qualitative analysis examples, and prompts employed in the LLMs, in further detail below.
\begin{figure}[b]
    \vspace{-1em}
    \centering
    \includegraphics[width=0.8\linewidth]{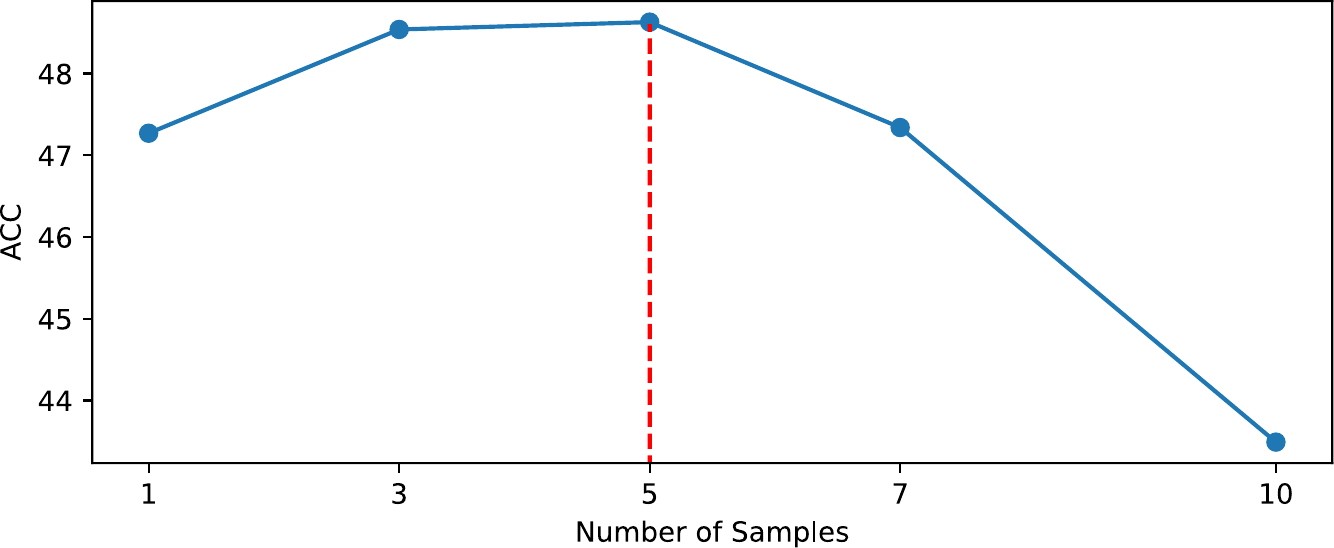}
    \vspace{-1em}\caption{Experimenting with different instruction sample sizes on the OK-VQA dataset. The vertical axis represents accuracy, while the horizontal axis denotes the number of instruction samples.}
    \label{fig:ablation_num_samples}
\end{figure}
\section{Number of Instruction Samples}
As outlined in approach section~\ref{sec:approach}, the LLM in our planner generated a set of $N$ instruction samples. We conducted an experiment using the OK-VQA dataset to determine the optimal number, which is detailed in Figure~\ref{fig:ablation_num_samples}. In our experiment, we tested various sample sizes, namely N = 1, 3, 5, 7, and 10. Upon examining the results presented in Figure~\ref{fig:ablation_num_samples}, we observed that employing 10 samples posed challenges for the RL agent's training and alignment with the complete set of actions. This challenge is particularly pronounced due to the modest size of the neural network composing the RL agent, essentially consisting of an MLP. As the number of instruction samples grows, and taking into account the limited dataset and the modest scale of the RL agent, incorporating a larger quantity of instruction samples becomes increasingly difficult for the small RL agent, consequently leading to a decline in performance. Conversely, with a very low number of actions (e.g., 1), the agent can request more instruction samples if it finds the current ones invalid, eventually obtaining a good sample. While a small number of actions makes it easier for a small RL agent to converge on meaningful decisions, it also increases the likelihood of rejecting all instruction samples compared to having five samples. Therefore, in terms of performance and efficiency, using five samples appears more promising, as depicted in the figure, which yielded an accuracy of 48.63\%.


\section{RL Agent Training}
As described in the approach section \ref{sec:approach}, the controller module comprises an RL agent implemented as an MLP utilizing the DQN algorithm \cite{mnih2013playing}. Below, we offer a more detailed explanation of the RL agent training process and its interaction with the environment, \memory{} and Meta Information, throughout.
\subsection{Embedding}
In the training phase, RL agent needs to interact with an environment, \memory~and Meta Information,\ which contains the current state and comprehensive information from the previous iteration. Meta information may encompass the system's own skills, its functionalities, and a description of a task (e.g. query). In \ours{},  we utilize a text-to-embedding template along with embedding models from the OpenAI API to obtain environmental information. This approach enables the learning of effective control policies from textual data within complex RL environments tailored for visual reasoning tasks. 
\subsection{Training process}
 We will detail the training process of the RL agent in this part. Initially, we configure a multi-layer perceptron with dimensions \{$1536$, $512$ and $N+1$\} with the use of \textit{text-embedding-3-small} model from the OpenAI API (N is the number of instruction samples, mentioned in section \ref{sec:approach}).
The learning start threshold is set to $1000$, indicating that the controller will make decisions randomly during the first $1000$ observation processes. We denote observation counts as $\omega$. The exploration epsilon is set at $0.2$, with an exploration epsilon decay rate of 0.02, and the epsilon decay interval is established at 200 steps. Therefore, the exploration threshold value is as:
\begin{equation}
 \mathcal{T} = \frac{0.2}{0.02 \times \frac{\omega}{200}}
\end{equation}
To enhance learning in visual tasks, the controller occasionally opts for random exploration over following the learned policy when a randomly generated value falls below the exploration threshold $\mathcal{T}$ within the range $[0,1]$.
As explained in the training phase in our approach, rewards for each action are calculated across a variety of environmental scenarios, with all corresponding rewards and situations stored in a reward buffer. We have optimized the batch size to 128, and the learning rate is established at $1 \times 10^{-4}$. During each weight update cycle, a batch of samples, matching the batch size, is randomly drawn from the reward buffer to update the MLP's weights. For further details on the DQN updating process employing stochastic gradient descent and reward buffer, please refer to \cite{mnih2013playing}.

\section{Ablation Study}
\uline{\textbf{ViperGPT.}}
In this experiment, we follow the original ViperGPT official GitHub repository for all datasets. Note, however, that ViperGPT uses Codex, which is deprecated. Therefore, in our experiment, we replace Codex with GPT-3.5 Turbo-0613.\\
\uline{\textbf{HYDRA-IR-S.}}
The RL agent has been integrated to ViperGPT, providing it with the ability to make decision on keeping or re-generating the instruction from LLM. This integration aims to enhance the model's decision-making capabilities by allowing it to learn optimal policies through trial and error. With this addition, the ViperGPT achieved 5.77\% improvement on its results as shown in Table~\ref{ablation} (row $2$).\\
\uline{\textbf{HYDRA-RL-IR.}}
In this experiment, unlike the ViperGPT model, we asked the LLM to generate more than one instruction sample. As shown in Table~\ref{ablation} (row $3$), the performance increased by 2.5\% in terms of accuracy, reaching 39.84\%.\\
\noindent\uline{\textbf{HYDRA-IR.}} In this experiment, we removed Incremental Reasoning, which means the model no longer processes information incrementally or adaptively over multiple steps. This removal likely impacts the model's ability to reason and solve complex tasks that require multi-step reasoning or context-dependent decision-making. Consequently, the accuracy decreased slightly to 45.98\% as shown in Table~\ref{ablation} (row $4$).\\
\noindent\uline{\textbf{HYDRA-RL-S.}}
In this experiment, we removed sampling, meaning the model's LLM only generates one instruction sample, and the RL agent has been eliminated from our framework. As shown in Table~\ref{ablation} (row $5$), the model benefited from this adjustment, achieving an accuracy of 41.08\%.\\
\noindent\uline{\textbf{HYDRA-S.}}
In this experiment, we removed sampling, meaning the model's LLM only generates one instruction sample. As shown in Table~\ref{ablation} (row $6$), the removal of sampling negatively impacts the model's performance.\\
\noindent\uline{\textbf{HYDRA-RL.}} Similar to the previous experiment, the RL agent has been eliminated from our framework. This removal removes the model's ability to learn from rewards and adjust its behavior accordingly, potentially limiting its capability to perform tasks that require adaptive decision-making or exploration of the environment. Despite this, the model still achieved an accuracy of 46.93\% as shown in Table~\ref{ablation} row $7$.

\section{Fail Rate Analysis}
To analyse model stability, following the protocol suggested by~\cite{gupta2023visual} we manually reviewed $\sim$ 100 samples per dataset and categorized error sources into four groups as shown in Fig.~\ref{fig:fail_rate}. When the LLM can not provided any valid instruction, HYDRA's performance suffers as the controller can not select good instructions. This error type is the most common in GQA. Moreover, code generator problems like calling non-existent APIs can also impact stability, as seen in RefCOCO datasets. Therefore, using a more powerful LLM, \eg GPT-4, can mitigate the impact of planner constraints and code generator issues and improve HYDRA's performance. Additionally, insufficient precision of the foundation models also leads to errors, as shown in OKVQA, indicating the need to employ SoTA foundation models. 

\begin{figure}[t!]
\begin{center}
\includegraphics[width=0.8\linewidth]{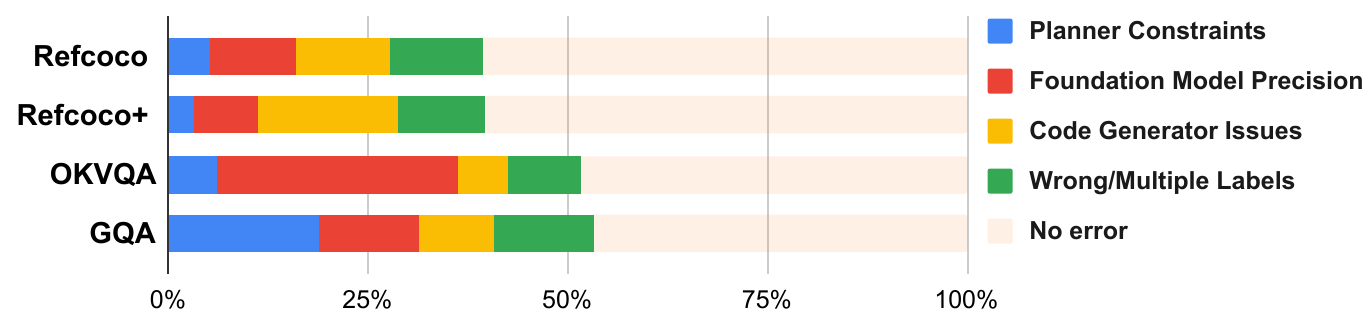} 
\end{center}
\caption{\ours~ Fail rate on each dataset.}
\label{fig:fail_rate}
\end{figure}

\section{Plug and Play in \ours}
HYDRA enhances visual reasoning tasks by leveraging its inherent capability to employ any foundation model as a VFM API. Intuitively, \ours's performance can be further improved by integrating the recent and larger foundation models, thus surpassing the performance of using a sole foundation model for the same task as shown in Table ~\ref{subtab:sotahydra}. This is because HYDRA has the ability to determine the appropriate API and utilize it at the correct step in the reasoning process. The experiment result Shows an improved accuracy of 62.5\% on the A-OKVQA dataset and 64.5\% on the GQA dataset when employing LLaVA-1.5. These results underscore the benefits of integrating cognitive agents for enhanced visual reasoning tasks and emphasize the advantages of the autonomy mechanism inherent in the compositional approach.

\begin{table}[t] 
\centering
\caption{
HYDRA Performance on GQA Dataset}
\label{subtab:sotahydra}\scalebox{1}{
{\fontsize{6}{10}\selectfont
\begin{tabular}{@{} c|c|c @{}} 
\toprule
Model & GQA ACC(\%) & A-OKVQA ACC(\%)\\ 
\midrule
BLIP2~\cite{li2023blip} & 45.5 & 53.7\\
\textbf{\ours}~w BLIP2 & \textbf{47.9} & \textbf{56.4}\\
\midrule
LLaVA1.5 (7B)~\cite{liu2024improved} & 62.0 & 61.6\\
\textbf{\ours}~w LLaVA1.5 (7B) & \textbf{64.5} & \textbf{62.5}\\
\bottomrule
\end{tabular}}}
\end{table}

\section{Textualizer Module Templates}
In the approach Section~\ref{sec:approach}, it is mentioned that when the perceptual output from the reasoner module is incomplete or unsuccessful, it undergoes conversion to textual format within \convertor{} module, as depicted in Figure~\ref{fig:framework}. The perceptual output from the reasoner, which may consist of bounding boxes, verifications, or captions, is transformed into a textual format using some templates. These templates are provided in Template~\hyperref[prompt:feedback]{C} which demonstrates the conversion of visually grounded fine-grained information into textual format. For instance, when the perception function \textit{find} is activated, the name of the target object is recorded in the detection results and the number of detected target objects is documented. Moreover, the bounding box coordinates for each target object are also recorded. The number of target objects and their locations, such as bounding box coordinates,  provide crucial information to the planner and controller for their subsequent actions.
Similarly, upon activating the perception function \textit{exists}, the system records the name of the object being checked, the name of the image, and the outcome of the check.
\begin{template}[title={Template C: Feedback Summarizer Examples}] \label{prompt:feedback}
\label{template}
\# \textbf{find}\\
Detection result: Only one \{object\_name\} has been detected in \{image\_name\}.\\
Detection result: \{num\} \{object\_name\} have been detected in \{image\_name\}.\\
Detection result: no \{object\_name\} has been detected.\\
Detected bounding box [x1,y1,x2,y2]: \{object\_name\}\_\{current\_img\_no\} in \{image\_name\} is \{bd\_box\_prediction\};\\
\# \textbf{existence}\\
The existence of \{object\_name\} in image patch \{image\_name\} is: \{exist\_result\}.\\
\# \textbf{verify}\\
The verification of \{category\} in \{image\_name\} is: \{verification\_result\}\\
\# \textbf{caption}\\
The caption for image patch \{image\_name\} is: \{caption\}.\\
\# \textbf{simple question answer}\\
The answer for image patch \{image\_name\} in response to the question `\{question\}' is: \{query\_answer\}\\
\# \textbf{depth calculation}\\
The median depth for image patch \{image\_name\} is: \{median\_depth\}\\
\# \textbf{LLM answer}\\
The obtained answer from LLM to the question, \{query\} with the additional context of \{context\} is: \{return\_answer\}\\
\# \textbf{sort}\\
The patches list has been sorted from left to right (horizontal). Now, the first patch in the list corresponds to the leftest position, while the last one corresponds to the rightest position.\\
\# \textbf{get middle patch}\\
The \{name\} is the middle one in the list.\\
\# \textbf{get the closest patch}\\
The \{name\} is the closest one to \{anchor\_name\}.\\
\# \textbf{get the farthest patch}\\
The \{name\} is the farthest one to \{anchor\_name\}.\\
\# \textbf{variables}\\
\{variable\_name\}: \{variable\_value\}
\end{template}

\section{More Qualitative Analysis Examples}
In this section, we offer a more qualitative analysis showcasing the output of each step in \ours{}, as illustrated in Figure \ref{fig:supp_example}. 
As depicted, the input image and query in the blue box are presented. Yellow boxes display the instructions step by step, while the green one shows their corresponding intermediate results.

\begin{figure}[t!]
\begin{center}
\includegraphics[width=\linewidth]{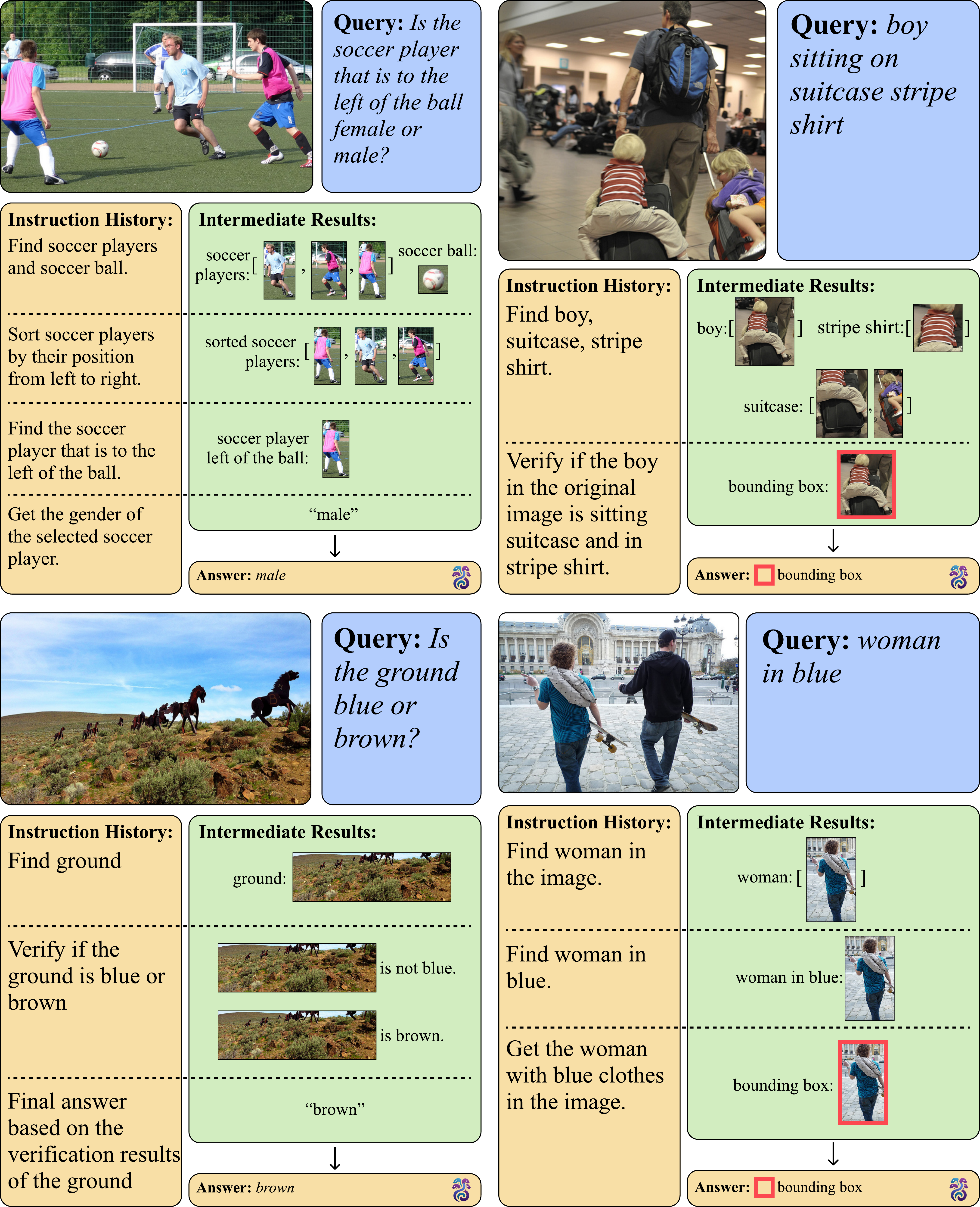} 
\end{center}
\vspace{-1em}\caption{More qualitative result examples from \ours{}.}
\label{fig:supp_example}
\end{figure}


\section{LLM's Prompts}
In \ours, we utilized LLMs in three distinct modes, as outlined in Section~\ref{sec:approach}: as an instruction sample generator in the planner, a code generator in the reasoner, and for summarizer in the textualizer module. For each mode, the prompt used is defined herein, and we provide detailed information on each prompt. Prompts~\hyperref[prompt:planner]{H.1}, \hyperref[prompt:code]{H.2}, and \hyperref[prompt:sum]{H.3} illustrate the abstract format of corresponding prompts for the instruction sample generator, code generator, and summarizer, respectively.
\begin{itemize}
    \item Prompt~\hyperref[prompt:planner]{H.1}: An instruction prompt is a crucial tool that informs the planner about the available perception skills, $\pi \in \Pi$ demonstrates the utilization of these skills, and describes how an instruction can leverage various skills for effective execution. Within the meta information, the skills are directly conveyed to the planner, informing it of \ours{}'s capabilities and guiding it towards generating appropriate subsequent instructions. Otherwise, the planner might generate instructions that cannot be executed.
    \item Prompt~\hyperref[prompt:code]{H.2}: A code prompt serves as a vital instrument, briefing the reasoner on the available perception skills along with templates for Python classes and functions. It includes a Python class and outlines several functions, showcasing the methodology for formulating Python code in response to the received instructions.
    \item Prompt~\hyperref[prompt:sum]{H.3}: A prompt functions as a guiding template, enabling the LLM to assess the adequacy of fine-grained information provided, based on the current state stored in \memory{}. If the detailed grounding information is adequate to address the query, the LLM will directly produce the answer, thus eliminating the need for further sequential responses.
\end{itemize}

\begin{prompt}[title={Prompt \thetcbcounter: Instruction Generation}]\label{prompt:planner}
[META\_INFO]

How to Use these Skills ideally:
[EXAMPLE]

Now the demonstration has ended. The following information are provided to you for recommending next-step instructions.

About Query: [QUERY\_TYPE]

Current Step: [CURRENT\_STEP\_NUM]

All Previously Taken Instruction:

[INSTRUCTION\_HISTORY]

Executed Python Code:\\
image\_patch = ImagePatch(image)

[CODE\_HISTORY]

Each variable details: [VARIABLE\_AND\_DETAILS]

Execution Feedback (Details of the known visual information in the image): 
[FEEDBACK\_HISTORY]

The question is `[QUERY]'

Please, provide [NUMBER\_OF\_SAMPLES] alternative instructions and associate each with a probability value indicating its likelihood of leading to the final answer. If available information is sufficient for answering question, please directly provide final answer as response.

Your response is here:
\end{prompt}

\begin{prompt}[title={Prompt \thetcbcounter: Code Generation}]\label{prompt:code}

[META\_INFO]

Provided Python Functions/Class:

[PYTHON\_API\_CODE]

Please only return valid python code:
If a Python variable is not found in the `Executed Python Code' section, it means that variable does not exist, and you cannot use any variable that has not been defined in the `Executed Python Code'.
        
[EXAMPLE]

Now the demonstration has ended. An instance (image\_patch = ImagePatch(image)) of the ImagePatch class is provided. 

Please translate only the `Current Instruction' into Python code. 
If the `Current Instruction' mentions the final process, assign the result to the variable named {final\_answer} for the concluding statement. If there is no mention of a final process in the current instruction, refrain from using {final\_answer}.
If a Python variable is not found in the `Executed Python Code' section, it means that variable does not exist, and you cannot use any variable that has not been defined in the `Executed Python Code'.
About Query: [QUERY\_TYPE]

Query: [QUERY]

Current Step: [CURRENT\_STEP\_NUM]

All Previously Taken Instruction: 

[INSTRUCTION\_HISTORY]

Executed Python Code:
image\_patch = ImagePatch(image)

[CODE\_HISTORY]

Each variable details:[VARIABLE\_AND\_DETAILS]

Execution Feedback (Details of the known visual information in the image): 
[FEEDBACK\_HISTORY]
   
Current Instruction: [CURRENT\_INSTRUCTION]

Generated Python Code for Current Instruction \-\- [CURRENT\_INSTRUCTION] here:
\end{prompt}

\begin{prompt}[title={Prompt \thetcbcounter: Summarizer}]\label{prompt:sum}
[META\_INFO]

Below is the information related to the question along with known visual details
About Question: 
[QUERY\_TYPE]

All Previously Taken Instruction: 

[INSTRUCTION\_HISTORY]

Executed Python Code:\\
image\_patch = ImagePatch(image)
[CODE\_HISTORY]

Each variable details: [VARIABLE\_AND\_DETAILS]
\\
Execution Feedback (Details of the known visual information in the image): 
[FEEDBACK\_HISTORY]
\\
The question is `[QUERY]'
\\
You need to base on details of the known visual information in the image to answer question. Respond concisely with key terms or names related to the question. 
Base your deductions solely on the Execution Feedback, which provides details of the known visual information in the image. Avoid making any random guesses if the available evidence does not sufficiently support your answer.
For example, When provided with limited information that only identifies an object as a fruit without further details, it's crucial to avoid making arbitrary guesses about the fruit's identity. Instead, the response should acknowledge the insufficiency of the data for a definitive identification.
If available information is insufficient for a definitive answer, reply with `continue'. 

Your answer is here:
\end{prompt}

\end{document}